\pdfoutput=1
\documentclass[11pt]{article}

\usepackage[]{acl}

\usepackage{times}
\usepackage{latexsym}
\usepackage[T1]{fontenc}
\usepackage[utf8]{inputenc}
\usepackage{microtype}
\usepackage{graphicx}
\usepackage{float}
\usepackage{graphics}
\usepackage[normalem]{ulem}
\usepackage{xspace}
\usepackage{mathtools}
\usepackage{bbm}
\usepackage{caption}
\usepackage{makecell}
\usepackage{multirow}

\newcommand{\draftcomment}[3]{{\textcolor{#3}{[#1]#2}}}

\newcommand{\inbal}[1]{\draftcomment{#1}{\textsc{Inbal}}{orange}}

\newcommand{\resolved}[1]{}
\newcommand{\com}[1]{}

\newcommand{\secref}[1]{Sec.~\ref{sec:#1}}
\newcommand{\appref}[1]{App.~\ref{app:#1}}

\newcommand{\figref}[1]{Fig.~\ref{fig:#1}}
\newcommand{\tabref}[1]{Tab.~\ref{tab:#1}}

\newcommand{\seen}[0]{\textit{seen}\xspace}
\newcommand{\unseen}[0]{\textit{unseen}\xspace}
\newcommand{\exploitation}[0]{\texttt{expl}\xspace}
\newcommand{\memorization}[0]{\texttt{mem}\xspace}
\newcommand{\Exploitation}[0]{\texttt{expl}\xspace}
\newcommand{\Memorization}[0]{\texttt{mem}\xspace}

\newcommand{\figwidth}[0]{0.45\textwidth}

\title{Data Contamination: From Memorization to Exploitation}

\author{Inbal Magar \quad\quad Roy Schwartz\\
  School of Computer Science and Engineering, The Hebrew University of Jerusalem, Israel \\
  \texttt{\{inbal.magar,roy.schwartz1\}@mail.huji.ac.il}}

\begin{document}
\maketitle

\begin{abstract}
Pretrained language models are typically trained on massive web-based datasets, which are often ``contaminated'' with downstream test sets. It is not clear to what extent models exploit the contaminated data for downstream tasks. We present a principled method to study this question. We pretrain BERT models on joint corpora of Wikipedia and labeled downstream datasets, and fine-tune them on the relevant task. 
Comparing performance between samples \seen and \unseen during pretraining enables us to define and quantify levels of memorization and exploitation.
Experiments with two models and three downstream tasks show that exploitation exists in some cases, but in others the models  memorize the contaminated data, but do not exploit it. We show that these two measures are affected by different factors such as the number of duplications of the contaminated data and the model size. Our results highlight the importance of analyzing massive web-scale datasets to verify that progress in NLP is obtained by better language understanding and not better data exploitation.
\end{abstract}

\begin{figure}[t]
  \centering
  \includegraphics[width=\figwidth]{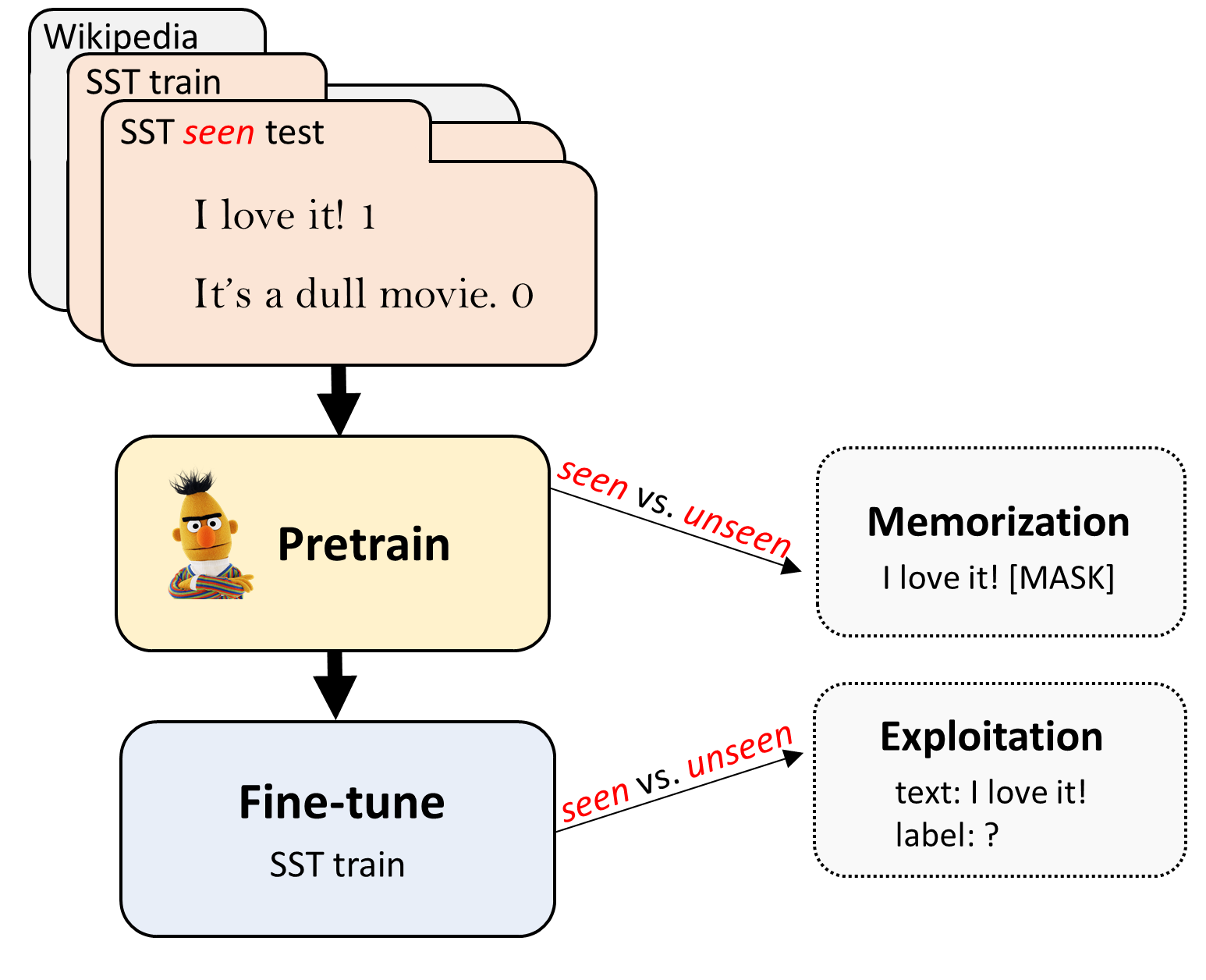}
  \caption{We pretrain BERT on Wikipedia along with both the labeled training and test sets (denoted \seen) of a downstream task (e.g., SST). Then, we fine-tune this model on the same training set for that task. We compare performance between samples \seen and \unseen during pretraining to quantify levels of memorization and exploitation of labels seen in pretraining. \com{Difference in performance indicates that the model exploits labels seen in pretraining.}}
  \label{fig:experient_setup}
\end{figure}

\section{Introduction}
Pretrained language models are getting bigger and so does their capacity to memorize data from the training phase \cite {carlini2021extracting}.
A rising concern regarding these models is ``data contamination''---when downstream test sets find their way into the pretrain corpus.    
For instance, \citet{dodge2021documenting}  examined five benchmarks and found that all had some level of contamination in the C4 corpus \cite{raffel2020exploring};  \citet{brown2020language} flagged over 90\% of GPT-3's downstream datasets as contaminated. Eliminating this phenomenon is challenging, as the size of the pretrain corpora makes studying them difficult \cite{Kreutzer:2021,Birhane:2021}, and even deduplication is not straightforward \cite{lee2021deduplicating}. 
It remains unclear to what extent data contamination affects downstream task performance.

This paper proposes a principled methodology to address this question in a controlled manner (\figref{experient_setup}). 
We focus on classification tasks, where instances appear in the pretrain corpus along with their gold labels.
We pretrain a masked language modeling (MLM) model (e.g., BERT; \citealp{devlin2019bert}) on a general corpus (e.g., Wikipedia) combined with labeled training and test samples (denoted \seen test samples) from a downstream task. We then fine-tune the model on the same labeled training set, and compare performance between \seen 
instances and \unseen ones, where the latter are unobserved in pretraining. We denote the difference between \seen and \unseen as \textit{exploitation}. We also define a measure of \textit{memorization} by comparing the MLM model's performance when predicting the masked label {for \seen and \unseen examples}. We study the connection between the two measures.

We apply our methodology to BERT-base and large, and experiment with three English text classification and NLI datasets. 
We show that exploitation exists, and is affected by various factors, such as the number of times the model encounters the contamination, the model size, and the amount of Wikipedia data. Interestingly, we show that memorization does not guarantee exploitation, and that factors such as the position of the contaminated data in the pretrain corpus and the learning rate affect these two measures. We conclude that labels seen during pretraining can be exploited in downstream tasks and urge others to continue developing better methods to study large-scale datasets.
As far as we know, our work is the first work to study the level of exploitation in a controlled manner.

\section{Our Method: Assessing the Effect of Contamination on Task Performance} 
\label{sec:method}

To study the effect of data contamination on downstream task performance, we take a controlled approach to identify and isolate factors that affect this phenomenon.
We assume that test instances appear in the pretrain corpus \textit{with their gold labels},\footnote{Our focus is on classification tasks, but our method can similarly be applied to other tasks, e.g., question answering.} and that the \textit{labeled} training data is also found in the pretrain corpus.\footnote{We recognize that these assumptions might not always hold; e.g., the data might appear unlabeled. Such cases, while interesting, are beyond the scope of this paper.} We describe our approach below.

We pretrain an MLM model on a general  corpus combined with a downstream task corpus, containing labeled training and test examples. We split the test set into two, adding one part to the pretrain corpus (denoted \seen), leaving the other unobserved during pretraining (\unseen).\resolved{ \inbal{need sentence to explain the choice with this contamination}}
For example, we add the following SST-2 instance  \cite{socher-etal-2013-recursive}:
\vspace{-20pt}
\begin{figure}[H]

    \centering{\texttt{I love it! 1 }\footnotemark}
    \captionsetup{labelformat=empty}
  \label{fig:sst_samples}
\end{figure}
\vspace{-10pt}

\footnotetext{One could imagine other formats, e.g., the label coming before (rather than after) the text. Preliminary experiments with this format showed very similar results.}

We then fine-tune the model on the \textit{same} labeled training set, and compare performance on the \seen and \unseen test sets. As both test sets are drawn randomly from the same distribution, differences in performance indicate that the model exploits the labeled examples observed during pretraining (\figref{experient_setup}).
This controlled manipulation allows us to define two measures of contamination: 

\paragraph{\Memorization}
is a simple measure of explicit memorization. We consider the MLM task of assigning the highest probability to the gold label (among the candidate label set); given the instance text (e.g., \texttt{I love it!~[MASK]}). \memorization is defined as the difference in MLM accuracy by the pretrained model (before fine-tuning) between \seen and \unseen.\footnote{Other definitions of memorization, such as relative log-perplexity of a sequence, have been proposed \cite{carlini2019secret, carlini2021extracting}. As we are interested in comparing the model's ability to predict the correct label, we use this strict measure.}
 
\memorization is inspired by recent work on factual probing, which uses cloze-style prompts to asses the amount of factual information a model encodes \cite{petroni2019language, zhong2021factual}. Similarly to these works, \memorization can be interpreted as lower bound on memorization of contaminated labels.

\paragraph{\Exploitation}
is a measure of exploitation: the difference in task performance between \seen and \unseen.

\memorization and \exploitation are complementary measures for the gains from data contamination; \memorization is measured after pretraining, and \exploitation after fine-tuning. As we wish to explore different factors that influence \exploitation, it is also interesting to see how they affect \memorization, particularly whether \memorization leads to \exploitation and whether \exploitation requires \memorization. Interestingly, our results indicate that these measures are not necessarily tied.

\paragraph{Pretraining design choices}
Simulating language model pretraining under an academic budget is not an easy task. To enable direct comparisons between different factors, we pretrain medium-sized models (BERT-\{base,large\}) on relatively small corpora (up to 600M tokens). We recognize that some of the results in this paper may not generalize to larger models, trained on more data.  However, as data contamination is a prominent problem, we believe it is important to study its effects under lab conditions. We hope to encourage other research groups to apply our method at larger scales.

\section{Which Factors Affect Exploitation?}
\label{sec:which_factors_affect_memorization}

We study the extent to which pretrained models can memorize and exploit labels of downstream tasks seen during pretraining, and the factors that affect this phenomenon. 
We start by examining how many times a model should see the contaminated data in order to be able to exploit it. 

We pretrain BERT-base on MLM using a combined corpus of English Wikipedia (60M tokens), and increasing numbers of SST-5 copies \cite{socher-etal-2013-recursive}. To facilitate the large number of experiments in this paper, we randomly downsample SST-5 to subsets of 1,000 training, \seen and \unseen instances. 
We train for one epoch, due to the practical difference between the number of times the task data \textit{appears} in the corpus and the number of times the model \textit{sees} it. For example, if a contaminated instance appears in the corpus once, but the model is trained for 50 epochs, then in practice the model encounters the contaminated instance 50 times during training. Further exploration of the difference between these two notions is found in \appref{exp_changing_copies}.
See \appref{expetimental_details} for experimental details. We describe our 
results below.

\label{sec:section_3}

\begin{figure}[t]
  \centering
  \includegraphics[width=\figwidth]{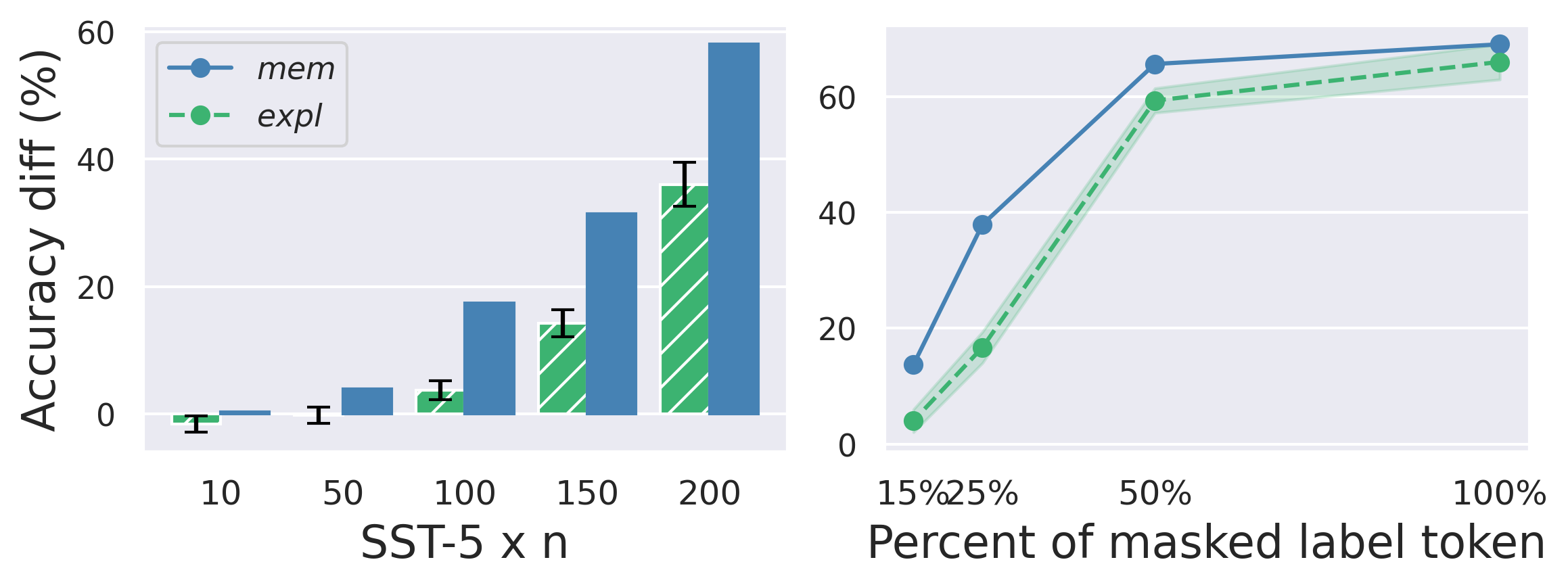}
  \caption{SST-5 \Memorization and \exploitation rise under different conditions. Left: increased number of data occurrences. Right: increased proportion of masking the label token. }
  \label{fig:combined_sst_size}
\end{figure}

\paragraph{Exploitation grows with contaminated data duplicates} 
\label{memorization_grows_with_occurences}

Both \memorization and \exploitation levels increase in proportion to the contaminated data, reaching 60\% \memorization and almost 40\% \exploitation when it appears 200 times (\figref{combined_sst_size}, left). 
This suggests a direct connection between both \memorization and \exploitation and the number of times the model sees these labels. This finding is consistent with several concurrent works, which show similar connections in GPT-based models.
These works study the impact of duplication of training sequence on regeneration of the sequence \cite{carlini2022quantifying, kandpal2022deduplicating}, and the effect on few-shot numerical reasoning \cite{razeghi2022impact}.
One explanation for this phenomenon is the increase in the expected number of times labels are masked during pretraining.\footnote{Following BERT, we mask each token with 15\% chance.} 
To check this, we pretrain BERT-base with 100 copies of SST-5 and varying probabilities of masking the label.
Our results (\figref{combined_sst_size}, right) show that the higher this probability, the higher \memorization and \exploitation values. 
These results motivate works on deduplication \cite{lee2021deduplicating}, especially considering that casual language models (e.g., GPT; \citealp{radford2019language}) are trained using next token prediction objective, and so every word in its turn is masked.

\begin{figure}[t]
  \centering
  \includegraphics[width=\figwidth]{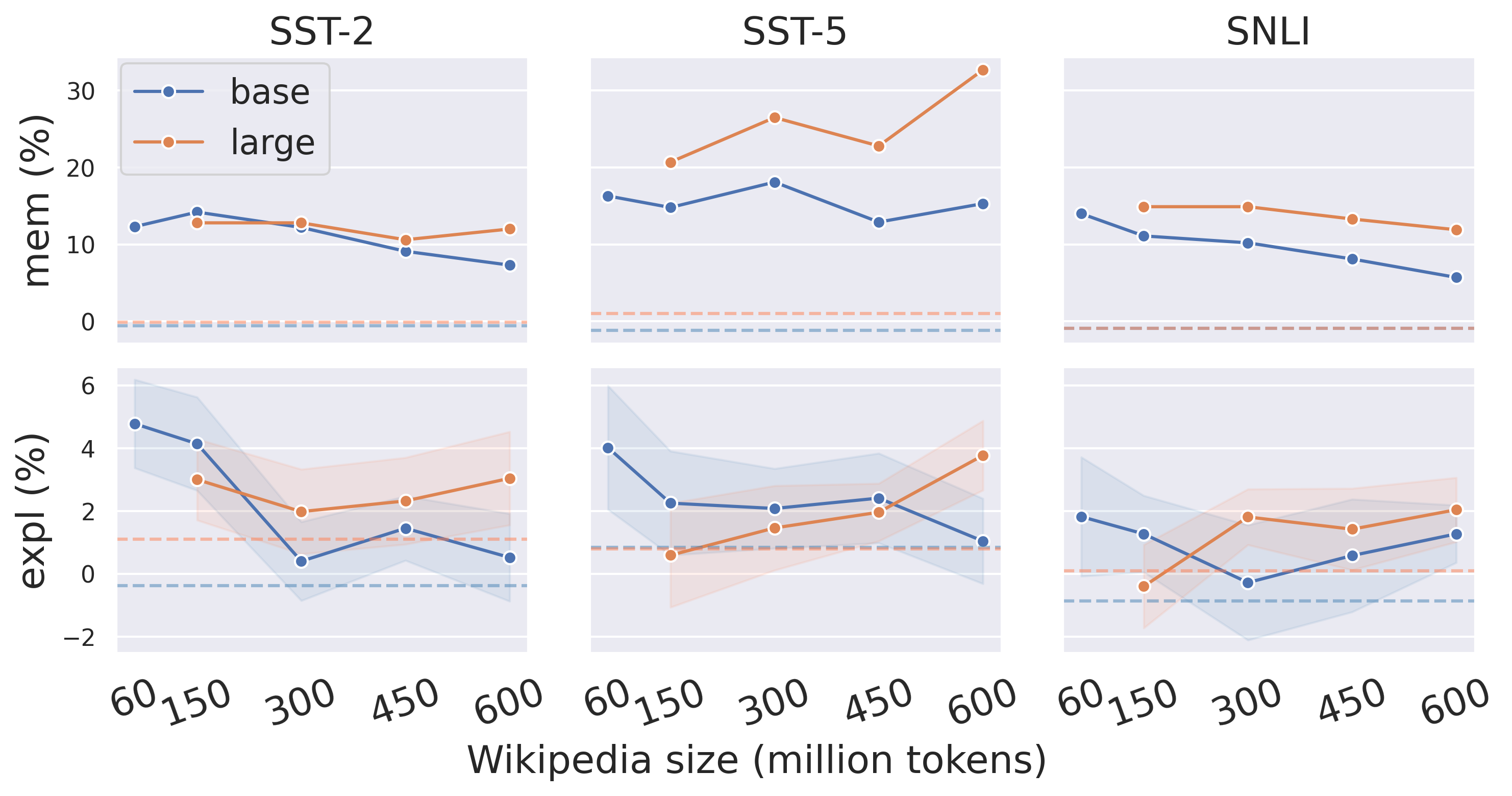}
  \caption{\Memorization and \exploitation of BERT-\{base,large\} on different tasks.
  We increase the size of clean data while fixing the amount of contaminated data.\footnotemark~\Exploitation values are averaged across ten random trials, shaded area corresponds to one SD. Dotted lines are \memorization/\exploitation baselines of BERT-\{base,large\} pretrained on uncontaminated data.}
  \label{fig:from_scratch_combined}
\end{figure}

In the following, we fix the number of contaminated data copies to 100 and modify other conditions---the size of the Wikipedia data and the model size (base/large). We also experiment with two additional downstream tasks: SST-2 and SNLI \cite{bowman2015large}. All other experimental details remain the same. \figref{from_scratch_combined} shows our results.

\paragraph{Memorization does not guarantee exploitation}

Perhaps the most interesting trend we observe is the connection between \memorization and \exploitation. Low \memorization values (10\% or less) lead to no \exploitation, but higher \memorization values do not guarantee \exploitation either. For example, training BERT-base with 600M Wikipedia tokens and SST-5 data leads to 15\% \memorization level, but less than 1\% \exploitation. These results indicate the \memorization alone is not a sufficient condition for \exploitation.

\footnotetext{Training of BERT-large models with 60M tokens did not converge, therefore they are not presented.}

\paragraph{Model and corpus sizes matter}
\label{model_size_matters}

Across all three datasets and almost all corpora sizes, \memorization levels of BERT-large are higher then BERT-base. This is consistent with \citet{carlini2021extracting}'s findings that larger models have larger memorization capacity. Also, we observe that \memorization levels (though not necessarily \exploitation) of SST-5 are consistently higher compared to the other datasets. This might be due to the fact that it is a harder dataset (a 5-label dataset, compared to 2/3 for the other two), with lower state-of-the-art results, so the model might have weaker ability to capture other features.

Much like memorization, exploitation is also affected by the size of the model, as well as the amount of additional clean data. We observe roughly the same trends for all three datasets, but not for the two models. For BERT-base, 2--6\% \exploitation is found for low amounts of clean data, but gradually decreases.  For BERT-large, the trend is opposite: \exploitation is observed starting 300M and continues to grow with the amount of external data, up to 2--4\%. This indicates that larger models benefit more from additional data.

We next explore other factors that affect \exploitation. Unless stated otherwise, we use BERT-base (60M Wikipedia tokens, 100 copies of SST-5).

\begin{figure}[!t]
  \centering
  \includegraphics[width=0.5\textwidth]{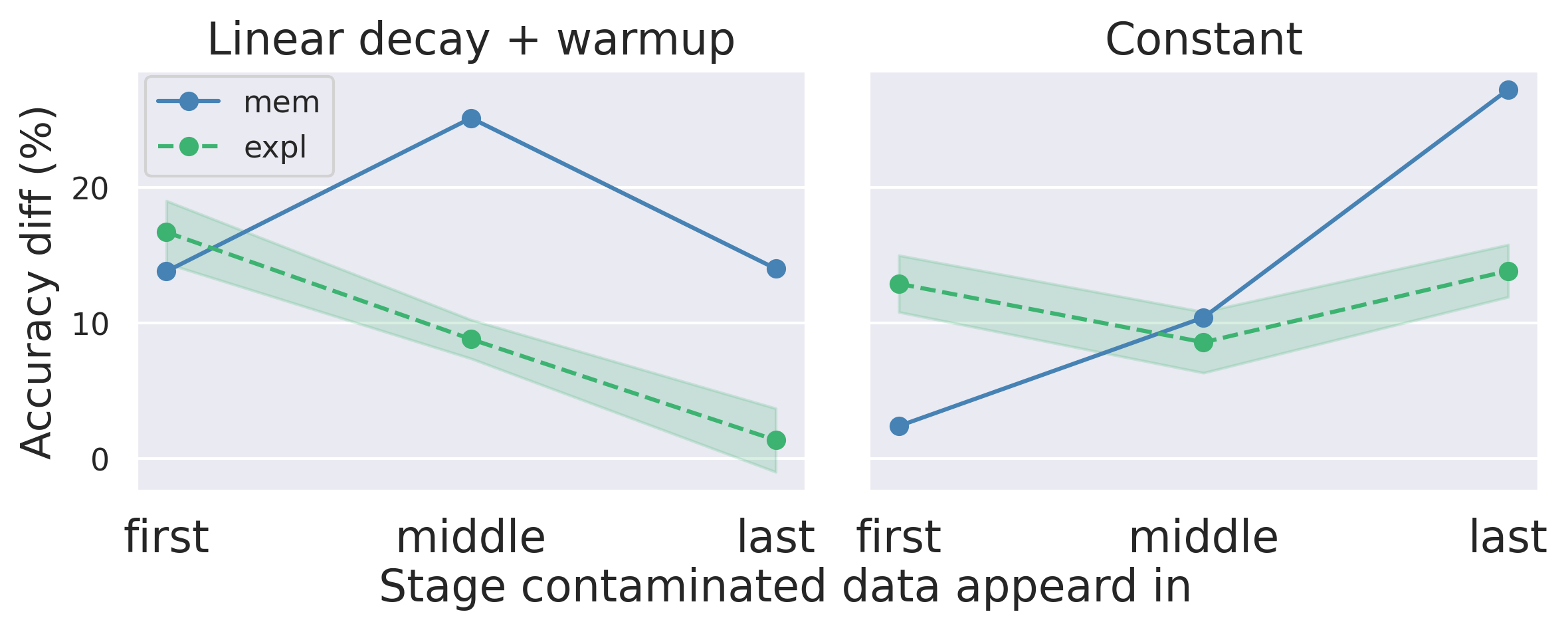}
  \caption{SST-5 \Memorization and \exploitation when contamination is inserted in different stages of pretraining, using a linear learning rate decay, and a constant learning rate.}
  \label{fig:combined_stage}
\end{figure}

\paragraph{Early contamination leads to high exploitation}
\label{by_stages}

Does the position of the contaminated data in the pretraining corpus matter? To answer this,
we pretrain the model while inserting contaminated data in different stages of pretraining: at the beginning (in the first third), the middle, or the end. Our results (\figref{combined_stage}, left) show that early contamination leads to high \exploitation (up to 17\%), which drops as contamination is introduced later.\footnote{Other datasets show a similar trend, see \figref{conatant_lr_sst2_snli}, \appref{stages}.} In contrast, the highest \memorization levels appear when contamination is inserted in the middle of the training. We also observe that in early contamination \memorization levels are \textit{lower} then \exploitation. This is rather surprising, since the model has certain level of memorization of the labels (as expressed by \exploitation), but it does not fully utilize these memories in the MLM task of \memorization. This suggests that in early contamination, the lower bound that  \memorization yields on memorization is not tight. The model might have an ``implicit'' memories of the labels, which are not translated to gains in the MLM task of predicting the gold label (\memorization).
Distinguishing between implicit and explicit memory of LMs is an important question for future work.

We note that different stages of training also yield different learning rates (LRs). In our experiments we follow BERT, using linear LR decay with warmup.

We might expect instances observed later, with lower LR, to have a smaller affect on the model's weights, thus less memorized. \figref{combined_stage} (left) indeed shows that late contamination leads to no \exploitation (though \memorization levels remain relatively high). To separate the LR from the contamination timing, we repeat that experiment with a constant LR of 2.77e-5 (midway of the linear decay). \figref{combined_stage} (right) shows that in the last stage, both measures increase compared to the LR decay policy. As the LR is constant, this indicates that both LR and contamination timing might affect label memorization.

\paragraph{Large batch size during pretraining  reduces exploitation}
\label{large_batch_reduce_memorization} 

\begin{figure}[t]
  \centering
  \includegraphics[width=\figwidth]{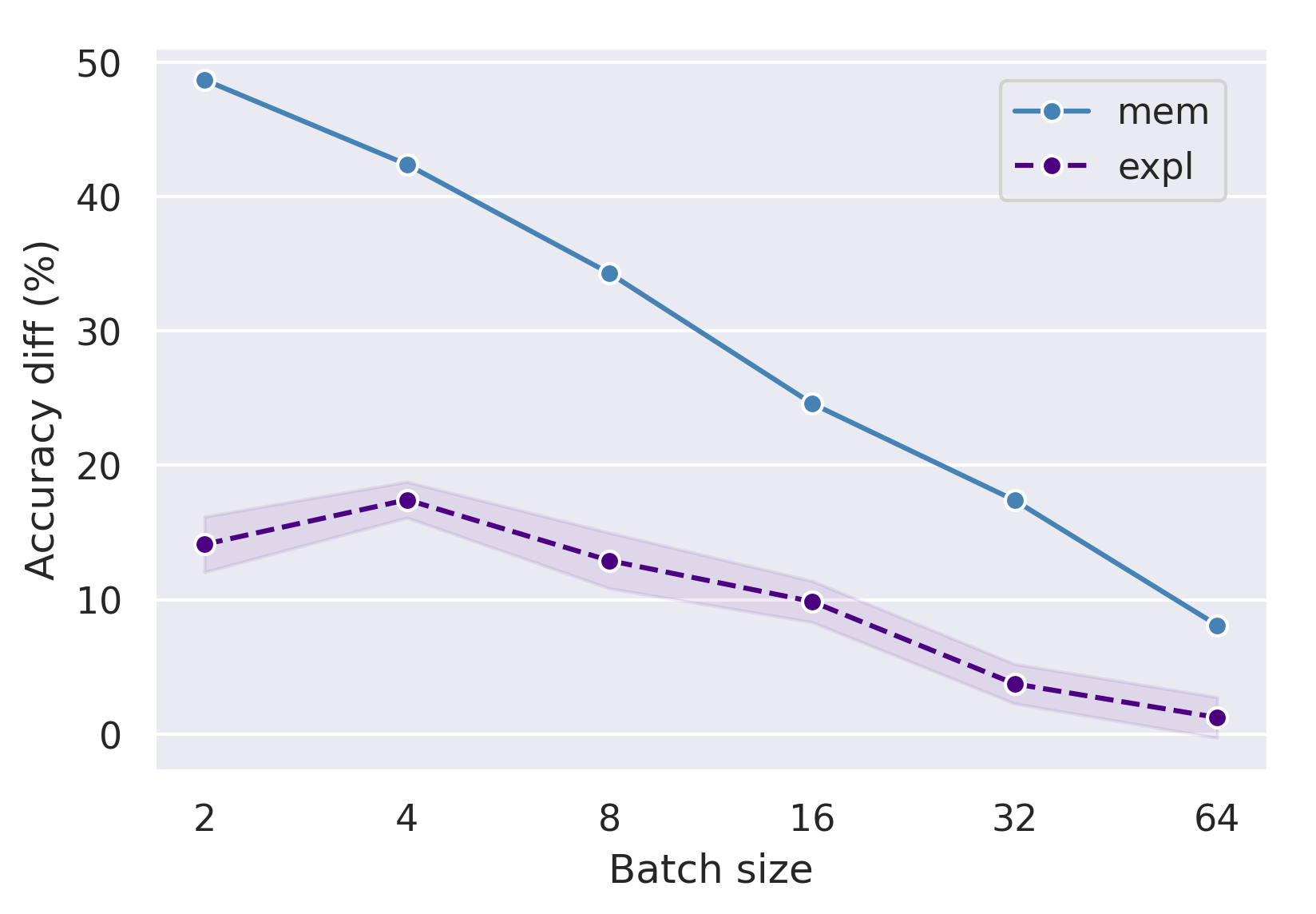}
  \caption{SST-5 \Memorization and \exploitation values drop as the pretraining batch size increases.}
  \label{fig:batch_size}
\end{figure}

Similar to learning rate, the batch size can also mediate the influence that each instance has on the models weights. We pretrain BERT-base several times with increasing batch sizes.\footnote{We update after each batch (no gradient accumulation).} Our experiments show that as we decrease the batch size, both measures increases (\figref{batch_size}). In the extreme case of batch size=2, \memorization reaches 49\%, and \exploitation reaches 14\%. This phenomenon might be explained by each training instance having a larger impact on the gradient updates with small batches.

\paragraph{A good initialization matters}
\citet{carlini2019secret} showed that memorization highly depends on the choice of hyperparameters. We observe a similar trend---\exploitation depends on the random seed used during fine-tuning. These results are also consistent with prior work that showed that fine-tuning performance is sensitive to the selection of the random seed \cite{dodge2020finetuning}.  Careful investigation reveals that some random seeds lead to good generalization, as observed by \unseen performance, while others lead to high exploitation: When considering the top three seeds (averaged across experiments) for \exploitation---two out of those seeds are also in the \textit{worst} three seeds for generalization. This indicates a tradeoff between generalization and exploitation. Future work will further study the connection between these concepts. To support such research, we publicly release our experimental results.\footnote{\url{https://github.com/schwartz-lab-NLP/data_contamination}}

\section{Related Work}
Memorization in language models has been extensively studied, but there is far less research on data contamination and the extent models exploit the contamination for downstream tasks. Most related to our work is \citet{brown2020language}'s post-hoc analysis of GPT-3's contamination. They showed that in some cases there was great difference in performance between `clean' and `contaminated' datasets, while in others negligible. However, they could not perform a controlled experiment due to the high costs of training their models. As far as we know, our work is the first work to study the level of exploitation in a controlled manner.

Several concurrent works explored related questions on memorization or utilization of training instances. These works mostly use GPT-based models. 
\citet{carlini2022quantifying} showed that memorization of language models grows with model size, training data duplicates, and the prompt length. They further found that masked language models memorize an order of magnitude less data compared to causal language model. This finding hints that exploitation levels might be even higher on the latter. \citet{kandpal2022deduplicating} showed that success of privacy attacks on large language models (as the one used in \citealp{carlini2021extracting}) is largely due to duplication in commonly used web-scraped training sets. Specifically, they found that the rate at which language models regenerate training sequences is superlinearly related to a duplication of the sequence in the corpus. Lastly, \citet{razeghi2022impact} examined the correlations between model performance on test instances and the frequency of terms from those instances in the pretraining data. They experimented with numerical deduction tasks and showed that models are consistently more accurate on instances whose terms are more prevalent.

\section{Discussion and Conclusion}
We presented a method for studying the extent to which data contamination affects downstream fine-tuning performance. Our method allows to quantify the explicit \textit{memorization} of labels from the pretraining phase and their \textit{exploitation} in fine-tuning.
 
Recent years have seen improvements in prompt-based methods for zero- and few-shot learning \cite{shin2020autoprompt,schick2021its,gu2021ppt}. These works argue that masked language models have an inherent capability to
perform classification tasks by reformulating them as fill-in-the-blanks problems. 

We have shown that given that the language model has seen the gold label, it is able to memorize and retrieve that label under some conditions.  
Prompt-tuning methods, which learn discrete prompts \cite{shin2020autoprompt} or continuous ones \cite{zhong2021factual}, might latch on to the memorized labels, and further amplify this phenomenon. This further highlights the importance of quantifying and mitigating data contamination.

\section*{Acknowledgements}
We wish to thank Yarden Shoham Tal, Michael Hassid, Yuval Reif, Deborah Elharar, Gabriel Stanovsky and Jesse Dodge for their feedback and insightful discussions.
We also thank the anonymous reviewers for their valuable comments. This work was supported in part by the Israel Science Foundation (grant no. 2045/21) and by a research gift from the Allen Institute for AI.

\bibliography{custom}
\bibliographystyle{acl_natbib}

\appendix

\section{Two Notions of ``Occurences''} \label{app:exp_changing_copies}
As noted in \secref{section_3}, the number of times an instance \textit{appears} in the corpus is a different notion than the number of times the model \textit{sees} it during training. The latter also takes into account the number of training epochs. For example, if an instance \textit{appears} in the corpus once, but the model is trained for 50 epochs, than practically the model \textit{sees} it 50 times. In the field on memorization and data contamination, it is mostly common to report the number of times an instance appears in the corpus \cite{carlini2021extracting, brown2020language}. However, the following experiments emphasizes the importance of accounting for the number of times a sample is seen. In the first experiment we fix the number of times the contamination  \textit{appears} in the corpus (10 copies), and change the number of times it is \textit{seen}. We do so by performing second-stage-pretraining \cite{gururangan2020dont, zhang2021inductive} on a combined corpus of Wikipedia and 10 copies of SST-5. We train one model for one epoch, and the other for 5 epochs. Results are shown in \tabref{10_appears}.
In the second experiment we fix the number of times the model \textit{sees} SST-5, and change the number of times it \textit{appears} in the corpus. We do so by performing second-stage-pretraining for one epoch on a combined corpus of Wikipedia and changing number of copies of SST-5. Results are shown in \tabref{50_seen}.

\begin{table}[]
    \centering
    \resizebox{0.6\columnwidth}{!}{%
    \begin{tabular}{ccc|c}
        \hline
        \textbf{\thead{epochs}} & \textbf{\thead{appears}} &
        \textbf{\thead{seen}}&
        \textbf{\thead{\exploitation}} \\
        \hline
1 & 10 & 10 &  2.07\% \\
5 & 10 & 50 &  6.87\%\\
        \hline
    \end{tabular}
    }
    \caption{\exploitation results of two models which were trained on corpus with 10 contaminated SST-5 \textit{appearances}.
    }
    \label{tab:10_appears}
\end{table}

\begin{table}[]
    \centering
    \resizebox{0.6\columnwidth}{!}{%
    \begin{tabular}{ccc|c}
        \hline
        \textbf{\thead{epochs}} & \textbf{\thead{appears}} &
        \textbf{\thead{seen}}&
        \textbf{\thead{\exploitation}} \\
        \hline
5 & 10 & 50 &  6.87\% \\
1 & 50 & 50 &  7.73\%\\
        \hline
    \end{tabular}
    }
    \caption{\exploitation results of two models which were \textit{saw} the contamination 50 times. 
    }
    \label{tab:50_seen}
\end{table}

We observe that \exploitation levels of the models which saw the contamination 50 times are rather similar. On the contrary, \exploitation levels of the model which saw the data 10 times is 5\% lower. These results indicate the number of times contamination is \textit{seen} during training have great influence on \exploitation. In the main experiments presented in this paper we train for one epoch in order to eliminate the difference between the two notion (\textit{appears} vs. \textit{seen}).

\section{Same Ratio, Different \Exploitation} \label{app:ratio}
\begin{figure}[H]
  \centering
  \includegraphics[width=\figwidth]{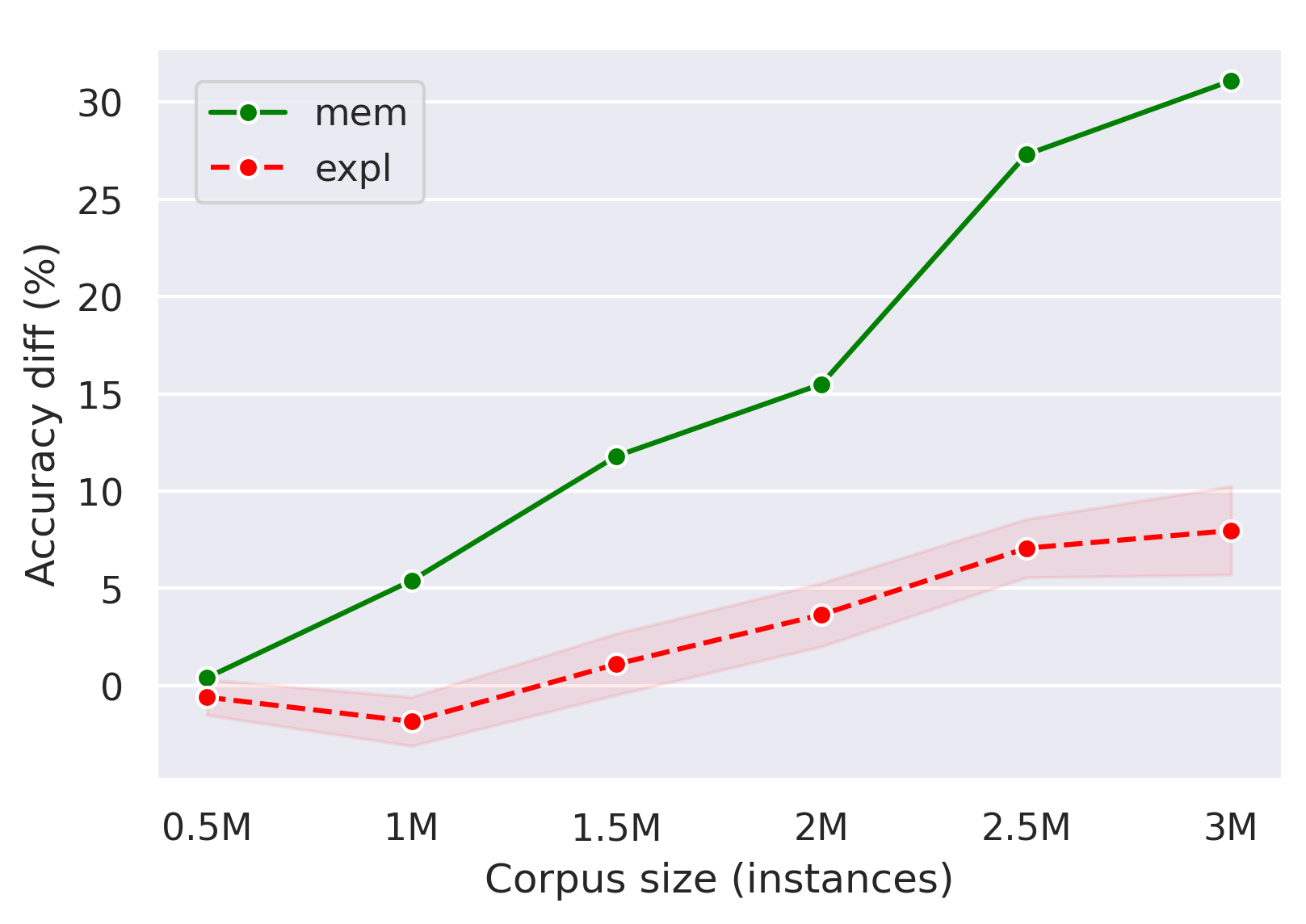}
  \caption{Keeping same ratio of 1:10 between contaminated data to total corpus by increasing both the number of SST-5 copies and the size of Wikipedia.}
  \label{fig:ratio}
\end{figure}

In \secref{which_factors_affect_memorization} we have seen the \exploitation and \memorization grows with the number of contamination occurrences in the corpus. One explanation for the results in is that the rising \textit{ratio} between the contaminated corpus and the full corpus leads to increased \memorization. We conduct experiments in which we keep the ratio between the two fixed while increasing their absolute sizes. We keep constant ratio of 1:10 between the number of instances (in Wikipedia set we consider lines as instances) in the datasets. To do so, we adjust both the size of Wikipedia and the duplications of SST-5 train and \seen test sets in the corpus. For example, to achieve total corpus sized 1M we use 9k instances from Wikipedia and 50 copies of SST-5 (which yields 1k samples). We focus on BERT-base and SST-5 task and follow the basic experiment setup and hyperparameters of our main experiments (\secref{which_factors_affect_memorization}). Our results (\figref{ratio}) show that this manipulation leads to increased \memorization, indicating the importance of the total number of occurrences of the task data.

\section{Position of Contamination Matters} \label{app:stages}
\begin{figure}[H]
  \centering
  \includegraphics[width=0.5\textwidth]{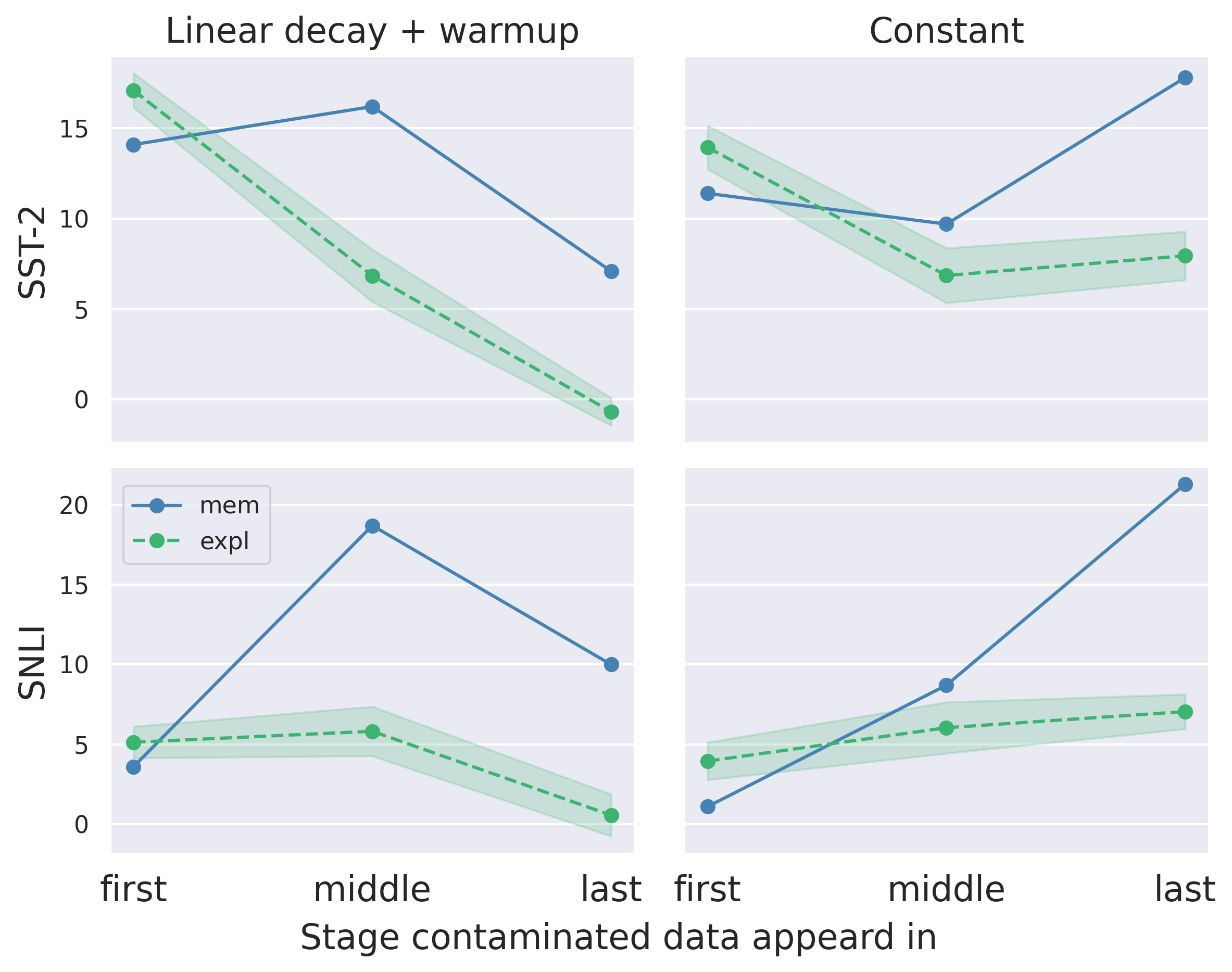}
  \caption{\Memorization and \exploitation when contamination is inserted in different stages of pretraining, using a linear learning rate decay, and a constant learning rate.}
  \label{fig:conatant_lr_sst2_snli}
\end{figure}

We pretrain BERT-base model while inserting contaminated data in different stages of pretraining. We discuss the experiment in \secref{which_factors_affect_memorization}. Results on SST-2 and SNLI can be found in \figref{conatant_lr_sst2_snli}.

\section{Experimental Details} \label{app:expetimental_details}

Originally, BERT model was trained on Masked Language Modelling (MLM) task and Next Sentence Prediction task (NSP; \citealp{devlin2019bert}). However, \citet{liu2019roberta} showed that removing the NSP loss does not impact the downstream task performance substantially. Therefore we pretrain both BERT models (-base and -large, both uncased) on the MLM task only.

\paragraph *{Wikipedia Data \label{data}}
We extracted and pre-processed the April 21' English Wikipedia dump. We used the wikiextractor tool \cite{Wikiextractor2015}. In order to measure the effect of contamination when contaminated data is shuffled across the pretraining corpus, we divided clean Wikipedia text into lines (instances which were originally separated by new line symbol).

\paragraph *{Experimental Details for \secref{which_factors_affect_memorization}}
All models were trained with the following standard procedure and hyperparameters. Specific experimental adjustments will be discussed later. We pretrained BERT models using huggingface's \cite{wolf-etal-2020-transformers} run\_mlm script for masked language modeling. We used heads sized 64 (calculated as: hidden dimension divided by the number of heads) with standard architecture as implemented in transformers library. We used a combined corpus of 60M tokens of Wikipedia along with 100 copies of the downstream corpus. 
Due to computational limitations, we limited the training sequences to 128 tokens. We pretrained for 1 epoch and used batch size of 32 to fit on 1 GPU. We trained with a learning rate of 5e-5. We apply linear learning rate warm up for the first 10\% steps of pretraining and linear learning rate decay for the rest. We fine-tune the models on 1,000 samples of the downstream corpora (SST-2, SST-5 and SNLI). 

We fine-tune for 3 epochs using batch size of 8. We use AdamW \cite{loshchilov2019decoupled} optimizer with learning rate of 2e-5 and default parameters: $\beta_1$ = 0.9, $\beta_2$ = 0.999, $\epsilon$ = 1e-6, with bias correction and without weight decay.
We average the results over ten random trials. As baselines we use pretrained BERT-base and BERT-large and fine-tune them as described above. Accuracy results on the \unseen test sets are shown in \tabref{abs_performance}.

\begin{table}[ht]
    \centering
    \resizebox{\columnwidth}{!}{
    \begin{tabular}{c|c|lllll|l}
        \hline
        \textbf{\thead{task}} & 
        \textbf{\thead{size}} & 
        \textbf{\thead{60M}} & 
        \textbf{\thead{150M}} & 
        \textbf{\thead{300M}} & 
        \textbf{\thead{450M}} & 
        \textbf{\thead{600M}} & 
        \textbf{\thead{baseline}} \\ 
        \hline
 
\multirow{2}{*}{SST-5} & base & 34.07 & 34.18 & 35.57 & 35.76 & 37.05 & 45.35 \\
& large &  & 33.76 & 32.93 & 34.3 & 37.1 & 48.28 \\
\hline
\multirow{2}{*}{SST-2} & base & 72.26 & 74.78 & 75.96 & 75.17 & 76.5 & 87.15\\
 & large &  & 70.49 & 73.5 & 73.76 & 73.85 & 89.29 \\
\hline
\multirow{2}{*}{SNLI} & base & 46.66 & 48.65 & 54.53 & 57.17 & 58.16 & 68\\
 & large &  & 47.58 & 49.61 & 55.53 & 59.05 & 67.11\\
\hline
    \end{tabular}
    }
    \caption{Accuracy of \unseen test set for main experiment in \secref{which_factors_affect_memorization}. 
    }
    \label{tab:abs_performance}
\end{table}

In the experiment of contamination in different stages of training, we divided the entire corpus (clean and contaminated) into 3 equal size sections, making sure that all the contaminated data appears entirely in one of those sections. We disabled the random sampler and shuffled each section individually. We refer to the sections as `first', `middle' and `last' according to the order they appear in training.  
All our experiments were conducted using the following GPUs: RTX 2080Ti, Quadro RTX 6000, A10 and A5000. 

\paragraph *{Experimental Details for \appref{exp_changing_copies}}
We conducted second-stage-pretraining by continuing to update BERT-base weights. We used batch size of 32 and learning rate of 5e-5. 
Learning rate scheduling, optimization and fine-tuning are the same as standard procedure described above.

\end{document}